\documentclass[english,francais]{article}
\usepackage{spconf,amsmath,graphicx}
\usepackage[utf8]{inputenc}  
\usepackage[T1]{fontenc}
\usepackage{float}
\usepackage{caption}
\usepackage{url}
\usepackage{gensymb}
\usepackage{booktabs}
\usepackage{dblfloatfix}
\usepackage{fixltx2e}
\usepackage{amssymb}
\usepackage{stmaryrd}
\usepackage{cite}
\usepackage{color}
\floatstyle{plaintop}
\restylefloat{table}

\title{Locally-adapted convolution-based super-resolution of irregularly-sampled ocean remote sensing data}

\name{Manuel López-Radcenco$^{\dagger}$ \quad Ronan Fablet$^{\dagger}$ \quad  Abdeldjalil Aïssa-El-Bey$^{\dagger}$ \quad Pierre Ailliot$^{\star}$\thanks{This work was supported by ANR (Agence Nationale de la Recherche, grant ANR-13-MONU-0014), Labex Cominlabs project SEACS and OSTST project MANATEE.}}
\address{$^{\dagger}$ IMT Atlantique, UMR CNRS 6285 Lab-STICC, Université Bretagne Loire\\Technopôle Brest-Iroise CS 83818, 29238 Brest Cedex 3, France \\
$^{\star}$ Laboratoire de Mathématiques de Bretagne Atlantique, UMR 6205, Université de Brest\\
6, Avenue Victor Le Gorgeu, B.P. 809, 29285 Brest Cedex, France}
\begin{document}
\ninept
\maketitle
\begin{abstract}
Super-resolution is a classical problem in image processing, with numerous applications to remote sensing image enhancement. Here, we address the super-resolution of irregularly-sampled remote sensing images. Using an optimal interpolation as the low-resolution reconstruction, we explore locally-adapted multimodal convolutional models and investigate different dictionary-based decompositions, namely based on principal component analysis (PCA), sparse priors and non-negativity constraints. We consider an application to the reconstruction of sea surface height (SSH) fields from two information sources, along-track altimeter data and sea surface temperature (SST) data. The reported experiments demonstrate the relevance of the proposed model, especially locally-adapted parametrizations with non-negativity constraints, to outperform optimally-interpolated reconstructions.
\end{abstract}
\begin{keywords}
Super-resolution, convolutional model, irregular sampling, dictionary-based decomposition, non-negativity
\end{keywords}
\section{Introduction}
\label{sec:intro}
Image super-resolution or upscaling is a classical problem in image processing \cite{SR_review,SR_single}. Super-resolution techniques also apply to remote sensing image enhancement problems \cite{SR_remote_sensing}. Contrary to the classical super-resolution setting, numerous satellite remote sensing applications do not only involve low-resolution images but also irregularly-sampled high-resolution information. The later may be due to specific sampling patterns, such as along-track narrow-swath satellite data, as well as to partial occlusions caused by weather conditions \cite{Fabletjstars2016,ElGheche2016}.
The availability of such partial high-resolution data supports locally-adapted super-resolution models, rather than models fully trained offline, with a view to accounting for the space-time variabilities of the monitored processes.\\
In this paper, we address such image super-resolution issues from irregularly-sampled high-resolution information. Following state-of-the-art super-resolution models \cite{ANR,Aplus,RB}, we consider locally-adapted convolution-based models. Our methodological contributions are two-fold: i) the proposed convolution-based models combine both a low-resolution image and a secondary image source, ii) we explore dictionary-based representations of the convolutional operators with different types of constraints, namely orthogonality, non-negativity and sparsity constraints \cite{NN_SR,sparse_SR}. Such dictionary-based representations and constraints are particularly appealing to resort to locally-adapted super-resolution models calibrated from a low number of high-resolution training data.\\
As case study, we apply the proposed framework to multi-source ocean remote sensing data, namely the reconstruction of high-resolution SSH (Sea Surface Height) images from satellite-derived along-track altimeter data, a high-resolution SST (Sea Surface Temperature) image and a low-resolution SSH image. We report numerical experiments, which demonstrate the relevance of the proposed super-resolution models, especially under non-negativity constraints, compared with optimally-interpolated SSH images.\\
The paper is organized as follows. In Section \ref{sec:model} we introduce the proposed super-resolution model along with the associated calibration schemes.
In Section \ref{sec:experiments}, we present the application to the reconstruction of satellite-derived SSH images and described experimental results.
Finally, we report concluding remarks and discuss future work in Section \ref{sec:conclusion}.

\section{Model formulation}
\label{sec:model}

\subsection{Problem statement}
We aim at reconstructing a series of high-resolution images $\{Y(t)\}_t$ at different times $\{t_1,....,t_T\}$ from the corresponding series of low-resolution images $\{Y_{LR}(t)\}_t$. In the considered application setting, we are also provided with:
\begin{itemize}
\item a complementary source of high-resolution images $\{X(t)\}_t$, which may depict some local or global correlation with $\{Y(t)\}_t$;
\item an irregularly-sampled dataset of high-resolution point-wise observations $\{\tilde{t}(k),\tilde{s}(k),\tilde{Y}(k)\}_k$, with $\tilde{t}(k)$, $\tilde{s}(k)$ and $\tilde{Y}(k)$ respectively the time, location and value of the $k^{th}$ high-resolution observation.
\end{itemize}
Figure \ref{fig:data} reports an example of the considered sampling patterns. We let the reader refer to Section 3 for the detailed description of the considered application to ocean remote data.\\
%
The reconstruction of high-resolution image $Y(t)$ given low-resolution image $Y_{LR}(t)$ is stated according to the following convolution-based model:
\begin{equation}
Y(t) = Y_{LR}(t) + H_{Y} * Y_{LR}(t) + H_{X} * X(t) + N(t)
\label{eq:model_conv}
\end{equation}
where $N$ is a space-time noise process. $H_{Y}$ (resp. $H_{X}$) is the two-dimensional impulse response of the $Y_{LR}$ (resp. $X$) component of the proposed convolutional model. $H_{Y}$ and $H_{X}$ are characterized by $(2W_p+1)\times(2W_p+1)$ discrete representations onto the considered high-resolution grid. Importantly, $H_{Y}$ and $H_{X}$ are space-and-time-varying operators and capture the space-time variabilities of $(Y,Y_{LR})$ and $(Y,X)$ relationships. This model can be regarded as a patch-based super-resolution approach where high-resolution image $Y$ at a given location is computed as a linear combination of $(2*W_p+1)\times (2*W_p+1)$ patches of images $X$ and $Y_{LR}$ centered at the same location. Parametrization $H_{X}=0$ clearly relates to regression-based super-resolution models \cite{Aplus,ANR}. 
%
%
\subsection{Unconstrained model calibration}
The calibration of model (\ref{eq:model_conv}) amounts to the estimation of 
the $(2W_p+1)\times(2W_p+1)$ matrix representations of operators $H_{Y}$ and $H_{X}$ at any space-time location. The availability of the irregularly-sampled dataset $\{\tilde{t}(k),\tilde{s}(k),\tilde{Y}(k)\}_k$ provides the means for this locally-adapted calibration. It may be noted that, in classical image super-resolution issue, such models are trained offline or involve nearest-neighbor techniques using a training dataset of joint low-resolution and high-resolution image patches \cite{Aplus,ANR}. Here, we proceed as follows. For a given space-time location $(t_0,s_0)$, we regard all data such that $ \tilde{t}(k) \in [t_0- D_t,t_0+D_t]$ and $\|\tilde{s}(k)-s_0\| \leq D_s$ as observations for model (\ref{eq:model_conv}) at location $(t_0,s_0)$. Parameters $D_t$ and $D_s$ state respectively the spatio-temporal extent of the considered neighborhood around location $(t_0,s_0)$.
Given the irregular sampling of the high-resolution dataset, no guarantees exist that sampling locations $\tilde{s}(k)$ will lie within the considered $X$/$Y_{LR}$ grid, and thus $(2W_p+1)\times(2W_p+1)$ high-resolution $X$ patches and low-resolution $Y_{LR}$ patches need to be interpolated around spatio-temporal locations $(\tilde{s}(k),\tilde{t}(k))$. Local impulse responses $H_{X}$ and $H_{Y}$ are then fitted 
by minimizing the mean square reconstruction error $\mathcal{E}\left(H_{X},H_{Y}\right)$ for the high-resolution detail $ dY=Y-Y_{LR}$ at irregularly-sampled dataset positions $(\tilde{s}(k),\tilde{t}(k))$:
\begin{equation}
\label{eq: mse}
\mathcal{E}\left(H_{X},H_{Y}\right)=\sum\limits_{k}\left\vert\left\vert d\tilde{Y}\left(k\right)-\widehat{d\tilde{Y}}\left(k\right)\right\vert\right\vert^2
\end{equation}
\begin{equation}
\begin{aligned}
\mbox{where}\quad\widehat{d\tilde{Y}}(k) = &H_{Y}*Y_{LR}\left(\tilde{t}\left(k\right),\tilde{s}\left(k\right)\right)+\\
&H_{X}*X\left(\tilde{t}\left(k\right),\tilde{s}\left(k\right)\right)
\end{aligned}
\end{equation}
Assuming the number of observations is high-enough, minimization (\ref{eq: mse}) resorts to a least-square estimation of operators $H_{Y}$ and $H_{X}$.

\subsection{Dictionary-based decompositions}
A critical aspect of the above least-square minimization is the number of available training data points and the underlying balance between locally-adapted and robust parametrizations. With a view to improving estimation robustness as well model interpretability, we explore dictionary-based decomposition approaches. They resort to the following decomposition of operators $H_{X}$ and $H_{Y}$:
\begin{equation}
H_{\{X,Y\}} = \sum_{k=1}^{K} \alpha_k D_k^{\{X,Y\}}
\label{eq:dict}
\end{equation}
where $D_k^{Y}$ (resp. $D_k^{X}$) is the k$^{th}$ component of the dictionary of operators for operator $H_{Y}$ (resp. $H_{X}$) and $\alpha_k$ is the k$^{th}$ scalar coefficient that states the decomposition of operator $H_{Y}$ (resp. $H_{X}$) onto dictionary element $D_k^{Y}$ (resp. $D_k^{X}$). It should be noted that a joint dictionary-based representation is considered in our study, so that decomposition coefficients $\alpha_k$ are shared by the two convolutional operators $H_{Y}$ and $H_{X}$.\\
%
%
Following classical dictionary-based settings \cite{dict}, we explore their applications to convolution operators. We investigate three different types of constraints for dictionary elements $\{D_k^{Y}\}$ and decomposition coefficients $\{\alpha_k\}$: namely orthogonality, sparsity and non-negativity constraints. The calibration of these dictionary-based settings first involve the estimation of dictionary elements $\{D_k^{Y}\}$ using training data. We here assume we are provided with a representative dataset of unconstrained estimates of operators $H_{Y}$ and $H_{X}$ from (\ref{eq: mse}), denoted by $\{H^n_{Y},H^n_{X}\}_n$. More precisely, the considered dictionary-based decompositions are as follows:
\begin{itemize}
\item {\bf Orthogonality constraint}: under this constraint, dictionary elements $\{D_k^{Y}\}$ form an orthonormal basis with no other constraints onto coefficients $\{\alpha_k\}$. This decomposition relates to the application of principal component analysis (PCA) \cite{pca} to dataset $\{H^n_{Y},H^n_{X}\}_n$. Given the trained dictionary, the estimation of decomposition coefficients $\{\alpha_k\}$ comes to the projection of the unconstrained operator estimates onto dictionary elements $\{D_k^{Y}\}$.
\item {\bf Sparsity constraint}: the sparse dictionary-based decomposition \cite{ksvd} resorts to complementing MSE criterion (\ref{eq: mse}) with the $L_1$ norm of coefficients $\{\alpha_k\}$. We apply a KSVD scheme to dataset $\{H^n_{Y},H^n_{X}\}_n$ to train dictionary elements $\{D_k^{Y}\}$. Given the trained dictionary, we proceed similarly to kSVD and use orthogonal matching pursuit \cite{omp} for the sparse estimation of decomposition coefficients $\{\alpha_k\}$ for any new unconstrained operator estimate.
\item {\bf Non-negativity constraint}: the non-negative dictionary-based decomposition constrains coefficients $\{\alpha_k\}$ to be non-negative. Given dataset $\{H^n_{Y},H^n_{X}\}_n$, the training of dictionary elements $\{D_k^{Y}\}$ resorts to the minimization of reconstruction error (\ref{eq: mse}) under non-negativity constraints for the decomposition coefficients. We exploit an iterative proximal operator-based algorithm \cite{proximal}. Given the trained dictionary, the estimation of decomposition coefficients $\{\alpha_k\}$ comes to a least-square estimation under non-negativity constraints.
\end{itemize}
\subsection{Locally-adapted dictionary-based convolutional models}
\begin{figure}[!tb]
\centering
\includegraphics[width=0.975\columnwidth]{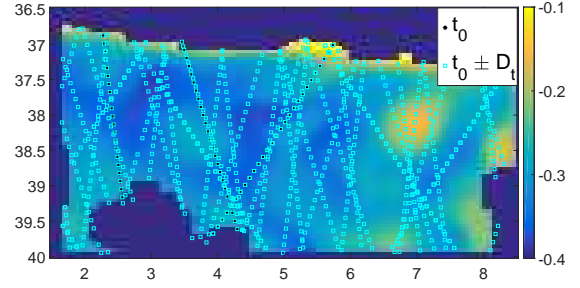}
\caption{Illustration of the irregular sampling of high-resolution observations associated with ocean remote sensing data: sea surface height image with the sampled along-track positions by satellite altimeters (cyan squares) in a $\pm 10$-day time window around April $20^{th}$, 2012.}
\label{fig:data}
\end{figure}
The application of the proposed dictionary-based decompositions to the super-resolution of irregularly-sampled high-resolution images involves the following main steps. For a given dictionary-based decomposition, we first train the associated dictionaries $\{D_k^{X},D_k^{Y}\}$. Considering the entire image time series, we proceed to the unconstrained estimation of operators $H_{X}$ and $H_{Y}$ from (\ref{eq: mse}) for a variety of spatio-temporal neighborhoods with given parameters $D_s^{Tr}$ and $D_t^{Tr}$. Parameters $D_s^{Tr}$ and $D_t^{Tr}$ are set such that the number of high-resolution observations is high enough to solve for least-square criterion (\ref{eq: mse}). We typically sample around 1500 neighborhoods to build a representative dataset of operators $H_{X}$ and $H_{Y}$.\\
Given the trained dictionaries, we proceed to the super-resolution of an image at a given date $t^*$ as follows. For any given spatial location $s^*$, we first estimate the associated decomposition coefficients $\{\alpha_k\}$ from the subset of high-resolution observations in a spatio-temporal neighborhood of space-time location $(t^*,s^*)$ with parameters $D_s^{SR}$ and $D_t^{SR}$. The later parameters typically define smaller spatio-temporal neighborhoods than training neighborhoods with parameters $D_s^{Tr}$ and $D_t^{Tr}$. As such, estimated coefficients $\{\alpha_k\}$ come to the projection of more local convolutional operators onto the subspace spanned by the estimated dictionaries, thus yielding a more locally-adapted model (\ref{eq:model_conv}). This calibrated model is then applied to the reconstruction of image $Y$ in a neighborhood of location $(t^*,s^*)$. To reduce the computational time, we perform this calibration of locally-adapted models for a regular subsampling of the image grid, typically $D_s^{SR}/2$, and use a spatial averaging of overlapping local reconstructions to obtain a single high-resolution reconstruction of image $Y$.
\section{Experiments} 
\label{sec:experiments}
\begin{figure}[!tb]
\centering
\includegraphics[width=0.96\columnwidth]{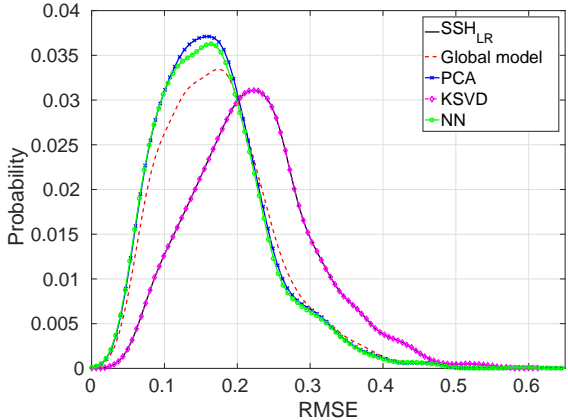}
\caption{Probability distribution for the relative root mean square reconstruction error (RMSE) for daily high-resolution SSH images $\{Y(t)\}_t$, for a global convolutional model and for locally-adapted decompositions of a global convolutional model using principal component analysis (PCA) \cite{pca}, KSVD \cite{ksvd} and non-negative decomposition (NN) and considering $K=10$ classes. The probability distribution of the RMSE for daily low-resolution SSH images $\{Y_{LR}(t)\}_t$ is given as reference (noted as $SSH_{LR}$).}
\label{fig:dist}
\end{figure}
\begin{table}[!tb]
\centering

\begin{tabular}{lccc}
\toprule
\multicolumn{1}{l}{} &\multicolumn{1}{l}{$K=2$} &\multicolumn{1}{l}{$K=5$} &\multicolumn{1}{c}{$K=10$}\\
\midrule
PCA 	
&{\bf 0.1807} 					&\underline{\bf 0.1734}			&\underline{0.1680}			\\     
KSVD 	
&{0.2228} 		&{0.2228}		&{0.2228} 		\\
NN      
&{\bf 0.1807} 	&\underline{\bf 0.1734} 	&\underline{\bf 0.1666} 	\\  
\midrule
\multicolumn{1}{l}{Global model} &\multicolumn{1}{l}{}&\multicolumn{1}{l}{} &\multicolumn{1}{c}{0.1755}\\
\multicolumn{1}{l}{$SSH_{LR}$}   &\multicolumn{1}{l}{}&\multicolumn{1}{l}{} &\multicolumn{1}{c}{0.2228}\\
\bottomrule
\end{tabular}

\caption{Relative root mean square reconstruction error (RMSE) for daily high-resolution SSH images $\{Y(t)\}_t$, for a global convolutional model and for locally-adapted decompositions of a global convolutional model using principal component analysis (PCA) \cite{pca}, KSVD \cite{ksvd} and non-negative decomposition (NN), considering $K=2$, $K=5$ and $K=10$ classes. The RMSE value for daily low-resolution SSH images $\{Y_{LR}(t)\}_t$ is given as reference (noted as $SSH_{LR}$). Best results for each number of classes $K$ considered are presented in bold. Results that outperform a global convolutional model are underlined.}
\label{tab:RMSE}
\end{table}
\begin{figure*}[!tb]
\centering
\includegraphics[width=0.92\textwidth]{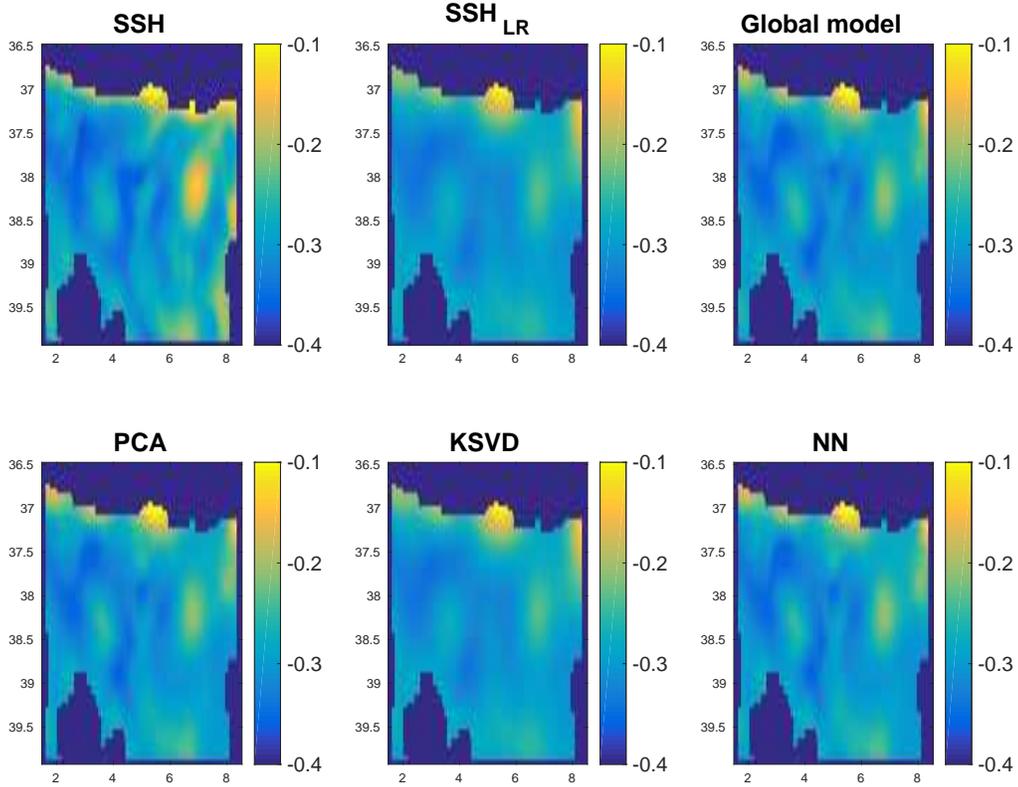}
\caption{{\bf High-resolution SSH image $Y$ reconstruction, April 20$^{th}$, 2012}: first row, from left to right, real high-resolution SSH image $Y$, low-resolution SSH image $Y_{LR}$ (noted as $SSH_{LR}$), reconstruction of high-resolution SSH image $Y$ using global convolutional model (\ref{eq:model_conv}); second row, reconstruction of high-resolution SSH image $Y$ using a 10-class locally-adapted decomposition (\ref{eq:dict}) of global convolutional model (\ref{eq:model_conv}) using, from left to right, principal component analysis (PCA) \cite{pca}, KSVD \cite{ksvd} and non-negative decomposition (NN).}
\label{fig:SSH}
\end{figure*}
As case study, we consider an application to ocean remote sensing data, more particularly to the reconstruction of sea-surface height (SSH) image time series from along-track altimeter data. Satellite altimeters are narrow-swath sensors such that high-resolution altimeter data is only acquired along the satellite track path \cite{pujol_mediterranean_2005}, resulting in an particularly scarce and irregular sampling of the ocean surface as illustrated in Fig.\ref{fig:data}. Interestingly, numerous studies have pointed out the potential contribution of high-resolution sea surface temperature (SST) images to the reconstruction of SSH images, as they share common geometrical patterns associated with the underlying upper ocean dynamics \cite{Lapeyre,Klein}. In addition, optimally-interpolated products \cite{pujol_mediterranean_2005} provide a low-resolution reconstruction of the SSH image. Overall, the reconstruction of high-resolution SSH image time series resorts to a super-resolution issue from irregularly-sampled high-resolution information as stated in Section \ref{sec:model}. It may be stressed that this case study involves a scaling factor of about 10 between the low-resolution and high-resolution data, which makes it particularly challenging compared with classical image super-resolution issues.\\
In our experiments, we exploit a ground-truthed dataset using an observing system simulation experiment for a case study region in the Western Mediterranean Sea ($36.5 \degree N$ to $40 \degree N$, $1.5 \degree E$ to $8.5 \degree E$). A high-resolution numerical simulation of the WMOP model \cite{juza_socib_2016} is used to generate daily high-resolution SSH images from 2009 to 2013 for a $1/20 \degree$ grid. The along-track dataset is simulated by sampling the SSH images at real along-track positions issued from from multiple altimetry missions in 2014 and 2015 (see Figure \ref{fig:data}). Given the simulated along-track dataset, optimally-interpolated SSH fields \cite{pujol_mediterranean_2005}, referred to as low-resolution SSH images $Y_{LR}$, are computed for a $1/8 \degree $ grid resolution. The calibration of the proposed convolutional operators is performed by considering $W_p=1$, which corresponds to $3 \times 3$ convolutional masks. We use the following parameter setting for spatio-temporal neighborhoods: 
$t_0\pm D_t$-day time windows with $D_t=10$, and $D_s\times D_s$ spatial neighborhoods with $D_s^{Tr}=7\degree$ for the training step and $D_s=2\degree$ for the locally-adapted calibration steps.\\
In Table \ref{tab:RMSE}, we report the average root mean square reconstruction error (RMSE) for daily high-resolution SSH images $\{Y(t)\}_t$, for a global convolutional model and for locally-adapted convolutional models, using principal component analysis (PCA) \cite{pca}, KSVD \cite{ksvd} and non-negative dictionary-based decomposition (NN) and considering $K=2$, $K=5$ and $K=10$ elements in the dictionaries. The reconstruction RMSE for daily low-resolution SSH images $\{Y_{LR}(t)\}_t$ (noted as $SSH_{LR}$) is given as reference.\\
From Table \ref{tab:RMSE}, locally-adapted convolutional models clearly outperform global models {for $K\geq 5$ (with the exception of the KSVD-based decomposition)}, which can be explained by the improved local adaptation to local spatio-temporal variabilities through locally-adapted decomposition coefficients. In this respect, the non-negative decomposition outperforms alternative approaches, with a maximum relative gain (with respect to optimally-interpolated low-resolution SSH images $\{Y_{LR}(t)\}_t$, at $K=10$) {of 25.22\% for NN, 24.60\% for PCA and 21.23\% for a global convolutional model.}\\
These results are further illustrated by the reconstruction of high-resolution SSH image $Y$ for sample date April 20$^{th}$, 2012 presented in Figure \ref{fig:SSH} and by the probability distributions of daily reconstruction root mean square error for high-resolution SSH images $\{Y(t)\}_t$, computed for the global convolutional model and for each one of the considered locally-adapted models with $K=10$, presented in Figure \ref{fig:dist}. Visually, the proposed super-resolution models clearly improve the reconstruction of finer-scale details compared to the low-resolution image. The model using non-negativity constraints seems to involve slightly sharper gradients compared with {the unconstrained model. The PCA-based model appears visually less relevant, while the KSVD-based model seems unable to exploit the high-resolution information sources to enhance the low-resolution altimetry field.} 

\section{Conclusion}
\label{sec:conclusion}
In this paper, we addressed the multimodal super-resolution of irregularly-sampled high-resolution images. This issue arises in a number of remote sensing applications, where several sensors associated with different regular and irregular sampling patterns may contribute to the reconstruction of a given high-resolution image. As a case study, we considered an application to the reconstruction of high-resolution sea surface height (SSH) images. From a methodological point of view, we complement previous convolution-based super-resolution models \cite{Aplus,RB} with the evaluation of different dictionary-based decompositions and the use of a complementary high-resolution image source. Dictionary-based decompositions are regarded as a means to better account for spatio-temporal variabilities through more locally-adapted model calibrations. Our numerical experiments support the selection  of non-negativity constraints to achieve a better local adaptation. They demonstrate the relevance of the proposed approach to achieve a better reconstruction of higher-resolution details, compared with the optimally-interpolated fields.\\
Future work includes non-local extensions of the proposed model to combine spatio-temporal and similarity-based neighborhoods as considered in regression-based super-resolution models \cite{Aplus,RB}. Non-linear dictionary-based decomposition seems particularly appealing to combine non-linear mapping, for instance CNN-based models \cite{SRCNN}, and locally-adapted models. As far as ocean remote sensing applications are considered, applying the proposed models to different sampling patterns, for instance along-track narrow-swath satellite data vs. wide-swath satellite data, appears to be of interest, the later possibly enabling the modeling of higher-order geometrical details.


\bibliographystyle{IEEEbib}
\bibliography{strings,refs}

\end{document}